\documentclass[twoside,11pt]{article}

\usepackage[preprint]{jmlr2e}
\usepackage{amsfonts}
\usepackage{amsmath}
\usepackage{mathrsfs}
\usepackage[svgnames]{xcolor}
\usepackage{graphicx}

\def\M{{\cal M}}
\def\R{{\mathbb{R}}}

\firstpageno{1}

\begin{document}
\title{Quadratic Multiform Separation: \\
A New Classification Model in Machine Learning}

\author{\name Ko-Hui Michael Fan\thanks{Corresponding author}
\email mkhfan@gmail.com\\
\addr AEM Technology Taiwan Corporation\\
Taipei, Taiwan, R.O.C.
\AND
\name Chih-Chung Chang \email ccchang@math.ntu.edu.tw\\
\addr Department of Mathematics\\
National Taiwan University\\
Taipei, Taiwan, R.O.C.
\AND
\name Kuang-Hsiao-Yin Kongguoluo \\
\addr AEM Technology Taiwan Corporation\\
Taipei, Taiwan, R.O.C.
}

\maketitle

\begin{abstract}
In this paper we present a new classification model in machine learning.
Our result is threefold: 1) The model produces
comparable predictive accuracy to that of most common classification models.
2) It runs significantly faster than
most common classification models.
3) It has the ability to 
identify a portion of unseen samples for
which class labels can be found with much higher predictive accuracy.
Currently there are several patents pending on
the proposed model \citep{Fan1,Fan2}.
\end{abstract}

\medskip
\begin{keywords}
quadratic multiform separation, loyalty extraction machine,
machine learning, supervised learning, statistical classification
\end{keywords}

\section{Introduction}
Machine learning has become one of the most important topics within
many scientific disciplines. What makes it attractive is that
it builds a model of a system solely based on observed input-output pairs
without referring to the system itself. Once the model ``template" is selected,
an algorithm is applied to modify model parameters
to match between the observed and predicted input-output pairs.
In this paper we consider the classification problem
and propose a novel model for its solution.
To our knowledge the proposed model does not resemble any existing ones.
Some of common classification models are
logistic regression, support vector machine,
naive bayes, decision tree, random forest, and neural network
(see, for example, \cite{Mitchell}, \cite{Nasteski}).

\medskip
The classification problem is defined as follows.
Let $\Omega\subset\R^p$ be a training set containing instances of $x$ from
input-output pair $(x,y)$ generated
by the {\it unknown} membership function $\mu:\Omega\to\M$.
Here $x\in\R^p$ is a vector, $y$ is a class label selected from the set
$$\M=\{1,\dots,m\}$$
and $m$ is the number of possible classes.
The problem is to find a classifier $\hat{\mu}(\cdot)$, serves as an approximation of
$\mu(\cdot)$, by equating the behavior of $\hat{\mu}(\cdot)$
to the samples in the training set. After a classifier is constructed,
a test set of unseen samples is used to evaluate its performance.
For a classification problem to be considered properly formulated,
the test set is assumed to possess
a similar sample distribution to that of the training set.

\medskip
For $i\in\M$, let $\Omega_i$ denote the member set defined by
$$\Omega_i=\{x\in\Omega:\mu(x)=i\}$$
Thus the collection of $\Omega_1,\ldots,\Omega_m$ forms
a partition of the training set $\Omega$.
The essence of the proposed model is quadratic multiform
separation (QMS), in which we seek pairwise
separation of the member sets using quadratic polynomials that satisfy
a certain cyclic conditions. We show that QMS is equivalent to
finding nonnegative quadratic polynomials $f_1,\ldots,f_m$ and defining
the classifier $\hat{\mu}$ by
\begin{equation}
\hat{\mu}(x)\in\{k:f_k(x)\le f_i(x)\text{~for~all~}i\} \label{s1-1}
\end{equation}
To this end, a QMS-specific loss function is introduced and a corresponding
optimization problem is formulated. We then propose
a simple yet very effective and efficient
algorithm, called coordinate perturbation method (CPM),
to minimize the loss function.
CPM is superior in speed because 
the method does not require full evaluation of the loss function
(not even just for once), nor it requires any gradient computation or line search of that sort.
With careful implementation, CPM appears to run significantly faster
than most of common classification models.

\medskip
We also move further into the notion of loyalty extraction machine (LEM).
The primary functionality of LEM is to
identify a specific portion of the test set for which
we will be able to find class labels with
much higher predictive accuracy.
LEM is related to the field of probabilistic classification and
has many advantages over non-probabilistic classifiers.

\medskip
Incidentally, in his master's thesis entitled
"Gradient-based quadratic multiform separation" \citep{Chang},
W.T. Chang studies the topic of designing
QMS classifiers using the Adam optimization algorithm \citep{Adam}.
The Adam optimization algorithm is an extension to stochastic
gradient descent that has recently seen broader adoption for
deep learning applications in computer vision and
natural language processing.
Chang's empirical numerical result shows that,
in terms of predictive accuracy, QMS performs
at least as good as most of common classification models.
Moreoever, QMS's superior performance is almost comparable to that
of gradient boosting algorithms which have won numerous machine
learning competitions.

\medskip
The remainder of the paper is organized as follows:
in Section 2 we present the theory of QMS,
in Section 3 we discuss LEM, in Section 4 we provide
numerical results of running QMS and LEM using the
Fashion MNIST dataset \citep{Fashion-MNIST},
and finally, Appendix A contains all the proofs.

\section{Quadratic Multiform Separation}
In this section we present the theory of QMS.
We start by introducing the concept of multiform separation.

\medskip\noindent
{\bf Definition 1.} {\it Multiform separation} is a mathematical model for classification.
It consists of $m$ piecewise continuous functions $f_1,\dots,f_m$,
which are constructed based on the training set $\Omega$.
Accordingly, a classifier
$\hat{\mu}:\Omega\to\M$ in the form of (\ref{s1-1}) can be realized. 
The functions $f_1,\ldots,f_m$ are called {\it separators}.
$\hfill\BlackBox$

\medskip
In practice, the functions $f_i$'s are chosen in a manner so that 
$\hat{\mu}(\cdot)$ serves as a good representation of $\mu(\cdot)$.
Just like every mathematical model in machine learning, selecting 
separators in multiform separation involves a trade-off between simplicity
and fidelity of the model. There is no doubt that the set of
piecewise continuous functions would be too large to handle.
We would like to add sufficient complexity
into the model to improve its realism, but maintain
the simplicity of the model
to make it easier to understand and analyze.
With that in mind, we propose to select separators
from the set of nonnegative quadratic polynomials, which 
we shall call {\it member functions}.

\medskip\noindent
{\bf Definition 2.} A polynomial $f:\R^p\to\R$ is said to be {\it nonnegative}
if $f(x)\ge 0$ for all $x\in\R^p$. 
\hfill$\BlackBox$

\medskip\noindent
{\bf Definition 3.} Let $q$ be a positive integer.
A function $f:\R^p\to\R$ is said to be a
{\it member funtion} (or $q$-{\it dimensional member function}) if it can be written as
$f(x)=\|Ax-b\|^2$ for some $A\in\R^{q\times p}$ and $b\in\R^q$,
where $\|\cdot\|$ denotes the Euclidean norm.
\hfill$\BlackBox$

\medskip\noindent
{\bf Lemma 1.} A quadratic polynomial $f:\R^p\to\R$ is nonnegative
if and only if it is a q-dimensional member function for
some $q\le p+1$.

\noindent
{\it Proof.} See Appendix A.
\hfill$\BlackBox$

\medskip\noindent
{\bf Definition 4.} {\it Quadratic multiform separation} is
a scenario of multiform separation where
the separators are selected among member functions.
$\hfill\BlackBox$

\subsection{Geometric Interpretation}
We provide below a geometric interpretation for QMS.
In this interpretation, we make use of the notion
of having Property A, and a result for the converse of a trivial statement:
the difference of two member functions is a quadratic polynomial.

\medskip\noindent
{\bf Definition 5.}
Given sets $\mathcal{S}_1,\mathcal{S}_2\subset\R^p$
and a quadratic polynomial $h:\R^p\rightarrow \R$,
the triplex $\{\mathcal{S}_1,\mathcal{S}_2, h\}$
is said to have {\it Property A} if
$h(x)<0$ for all $x\in\mathcal{S}_1$ and $h(x)>0$
for all $x\in\mathcal{S}_2$.
\hfill$\BlackBox$

\medskip\noindent
In other words, Property A asserts that
the sets $\mathcal{S}_1$ and $\mathcal{S}_2$ reside
on two different sides of the manifold defined by $h(x)=0$,
or the function $h$ separates the sets
$\mathcal{S}_1$ and $\mathcal{S}_2$.

\medskip\noindent
{\bf Lemma 2.} Every quadratic polynomial can be written
as the difference of two member functions. 

\noindent
{\it Proof.} See Appendix A.
\hfill$\BlackBox$

\medskip
Let us first consider the case when only two
classes in the classification problem, that is, $m=2$.
It is easy to see that in order to have a classifier $\hat{\mu}(\cdot)$
match perfectly the membership function $\mu(\cdot)$,
the task of QMS is essentially tailored to look for
member functions $f_1$ and $f_2$ in such a way that
\begin{eqnarray}
&& f_1(x)<f_2(x){\rm~for~all~}x\in\Omega_1,{\rm~and~}\label{s21-1} \\
&& f_2(x)<f_1(x){\rm~for~all~}x\in\Omega_2\label{s21-2}.
\end{eqnarray}
Suppose the conditions (\ref{s21-1}) and (\ref{s21-2}) hold for some
member functions $f_1$ and $f_2$.
It is obvious that $f_1-f_2$ is a quadratic polynomial and
the triplex $\{\Omega_1,\Omega_2,f_1-f_2\}$
has Property A. On the other hand, suppose that
the triplex $\{\Omega_1,\Omega_2,h\}$ has Property A
for some quadratic polynomial $h$. In view of Lemma 2, $h$
can be written as $h=f_1-f_2$ for some member functions $f_1$ and $f_2$.
Then it is also obvious that
the conditions (\ref{s21-1}) and (\ref{s21-2}) hold for $f_1$ and $f_2$.
This leads to an interesting interpretation of QMS:
for the case when $m=2$, QMS amounts to separating $\Omega_1$ and $\Omega_2$ by a
quadratic polynomial.
We assume that this quadratic polynomial $h$ exists
and it has been found by some method. Then the following simple procedure leads to a
perfect classifier: obtain member functions $f_1$ and $f_2$ so that $h=f_1-f_2$, and define
$\hat{\mu}(\cdot)$ by (\ref{s1-1}). The existence of a separating quadratic
polynomial ensures the existence of a perfect classifier. In reality, however,
such an ideal situation may not be true and
thus one might alternatively
seek an approximation of separation defined in terms of some optimal sense.

\medskip
Are we able to extend the above interpretation for the general $m$?
The answer is yes but it is more nuanced.
Consider the case when $m=3$.
The task of QMS to achieve a perfect
classifier in this case is essentially tailored to look for
member functions $f_1,f_2$, and $f_3$ to satisfy the conditions
\begin{eqnarray}
&& f_1(x)<f_2(x),~f_1(x)<f_3(x){\rm~for~all~}x\in\Omega_1,{\rm~and~}\label{s21-3} \\
&& f_2(x)<f_1(x),~f_2(x)<f_3(x){\rm~for~all~}x\in\Omega_2,{\rm~and~}\label{s21-4} \\
&& f_3(x)<f_1(x),~f_3(x)<f_2(x){\rm~for~all~}x\in\Omega_3\label{s21-5}
\end{eqnarray}
We assume the conditions (\ref{s21-3})-(\ref{s21-5}) hold for some
member functions $f_1,f_2$ and $f_3$, and
define the quadratic polynomials as $h_{12}=f_1-f_2$, $h_{23}=f_2-f_3$ and
$h_{31}=f_3-f_1$. Then it is easy to see that the following conditions hold.
\begin{eqnarray}
&& \{\Omega_1,\Omega_2,h_{12}\}{\rm~has~Property~A}\label{s21-6} \\
&& \{\Omega_2,\Omega_3,h_{23}\}{\rm~has~Property~A}\label{s21-7} \\
&& \{\Omega_3,\Omega_1,h_{31}\}{\rm~has~Property~A}\label{s21-8} \\
&& h_{12}+h_{23}+h_{31}=0 \label{s21-9}
\end{eqnarray}
Therefore it is very straightforward to derive from member functions $f_i$'s to
the corresponding quadratic polynomial $h_{ij}$'s.
The derivation for going the other direction
is less obvious. Lemma 3 asserts its feasibility. We will omit the proof here
since it is a special case of Theorem 1 (see below).

\medskip\noindent
{\bf Lemma 3.} Suppose
the conditions (\ref{s21-6})-(\ref{s21-9}) hold for some
quadratic polynomials $h_{12},h_{23}$ and $h_{31}$.
Then there exist member functions $f_1,f_2$ and $f_3$ so that
$h_{12}=f_1-f_2$, $h_{23}=f_2-f_3$,
$h_{31}=f_3-f_1$ and the conditions (\ref{s21-3})-(\ref{s21-5}) hold.
\hfill$\BlackBox$

\medskip\noindent
Lemma 3 reveals a somewhat surprising result. For the case when $m=3$, QMS is not
equivalent to {\it independently} solving
pairwise separation of the sets
$\Omega_1,\Omega_2$ and $\Omega_3$ by quadratic polynomials. Instead, Lemma 3 says
those quadratic polynomials have to satisfy
one additional constraint which is sum of them is equal to zero.

\medskip
Now we turn the discussion to the general case.
Theorem 1 below, which extends the result in Lemma 3, 
grants a geometric interpretation of QMS.
In addition, with the quadratic polynomials in place, explicit expressions
for constructing member functions are provided in its proof.
The geometric interpretation is stated as follows:
QMS is a mathematical model for classification in which one seeks pairwise
separation of the member sets using quadratic polynomials that satisfy
a certain cyclic conditions.

\medskip\noindent
{\bf Theorem 1.} Suppose there exist quadratic polynomials $h_{ij}$, $i,j\in\M$ such
that (i) the triplex $\{\Omega_i,\Omega_j,h_{ij}\}$ has Property A
for all $i,j\in\M, i\neq j$
and (ii) the cyclic condition
$$h_{ij}+h_{jk}+h_{ki}=0$$
holds for all $i,j,k\in\M$. Then there exist member functions $f_1,\dots,f_m$ such that,
for all $i\in\M$, the following condition holds
$$f_i(x)<f_j(x) {\rm~for~all~} x\in\Omega_i {\rm~and~for~all~} j\in\M,j\neq i$$

\noindent
{\it Proof.} See Appendix A.
\hfill$\BlackBox$

\subsection{Loss Function}
At this point we have completed the analysis stage 
of the classification problem.
In that stage, we employ a mathematical model and scientific
principles to help us predict the design results.
The next step is the optimization stage where a 
systematic process using design constraints
and criteria takes place to locate the optimal 
member functions $f_1,\ldots,f_m$.
To facilitate the process, a loss function must be defined.

\medskip
A loss function maps the member functions
$f_1,\ldots,f_m$ onto a real number intuitively representing some ``cost".
Let us first consider part of the loss function that is associated
with a particular sample, say, $x\in\Omega_1$.
This part of the loss function should resemble a degree of violation for the inequalities
$$f_1(x)<f_j(x){\rm~for~all~}j=2,\ldots,m$$
Therefore a simple candidate for its choice could be
\begin{equation}
\sum_{j=2}^m\max\left\{-\epsilon,f_1(x)-f_j(x)\right\} \label{s22-1}
\end{equation}
where $\epsilon$ is a small positive number.
However, (\ref{s22-1}) is not numerically sound
since it will depend on the problem scaling.
To be specific, we see that multiplying a positive
constant to all $f_i$'s will change (\ref{s22-1}) but it should not be doing so.
Instead, the following choice seems to be a better one.
\begin{equation}
\sum_{j=2}^m\max\left\{\alpha,\frac{f_1(x)}{f_j(x)}\right\} \label{s22-2}
\end{equation}
where $\alpha\in[0,1)$. The use of
$\alpha$ in (\ref{s22-2}) is to remove the constraint $f_1(x)<f_j(x)$
during optimization as long as $f_1(x)$ stays sufficiently smaller than $f_j(x)$.

\medskip
Now we include all $x$ from the training set $\Omega$ and formally present the definition.
Select $\alpha\in[0,1)$ and
define the loss function $\phi$ by
\begin{equation}
\phi=\sum_{i\in\M}\phi_i \label{s22-3}
\end{equation}
where, for $i\in\M$, $\phi_i$ denotes
\begin{equation}
\phi_i=\sum_{x\in\Omega_i}~\sum_{j\in\M,j\ne i}
\max\left\{\alpha,\frac{f_i(x)}{f_j(x)}\right\} \label{s22-4}
\end{equation}

\subsection{Coordinate Perturbation Method}
In this section, we propose a simple yet very effective and efficient
algorithm for minimizing the loss function defined in (\ref{s22-3}) and (\ref{s22-4}).
The algorithm works as follows. It successively minimizes
along each and every coordinate, one at a time, to find a descenting point.
Along each coordinate, the algorithm considers at most two nearby points,
one along each of two opposite directions. It moves to
a nearby descenting point or otherwise does nothing if both
nearby points do not produce a lower value for the loss function.
The algorithm then switches to the next coordinate and continues.
We shall call this algorithm the {\it coordinate perturbation method} (CPM).

\medskip
We point out that CPM does not require full evaluation of the loss function
$\phi$ (not even just for once), nor it requires any gradient
computation or line search of that sort.
With careful implementation, CPM appears to run significantly faster than most of common
classification models. In Section 4, among other things, 
we will report a timing result of CPM using the Fashion MNIST dataset.

\medskip
In closing this section, we would like to
mention one more thing. Our implementation of CPM uses only
single-thread/no-GPU computer resource.
In spite of its exceptional speed performance, 
it is not clear how to take full advantage of the multi-cpu/multi-thread/GPU
features that exist in today's computer hardware. This is currently under investigation. 

\section{Loyalty Extraction Machine}
In machine learning, an ensemble method uses
multiple classifiers to obtain better
predictive performance than could be obtained from any of the constituent
classifiers alone. In this section, we also consider multiple classifiers
created by altering the definition of the loss function.
Even though our method improves overall 
predictive performance, it is not our goal to do so.
Instead, among other things, we would like to
identify a specific portion of the test set for which 
class labels can be found with much higher predictive accuracy.
In order to achieve this goal, a scheme (or a machine) is
developed to divide the test set according to the so-called
loyalty type of its samples. There are three loyalty
types defined here: strong, weak, and normal. A finer partition of the test set
is readily applicable with the expenses of complexity and computation time.
As a result, the test set turns into the union of three
disjoint subsets, each corresponds to a different loyalty type.
We shall call this process the {\it loyalty extraction machine} (LEM).
The loyalty type is an estimate of how confident a sample adheres to its
predicted class. This estimate of course is based on the statistical information
learned from the training set.

\medskip
LEM is related to the field of probabilistic classification.
A probabilistic algorithm computes a probability of
the sample being a member of each of the possible classes.
The best class is normally then selected as the one
with the highest probability. Similar to the probabilistic
algorithms, LEM has the following advantages over
non-probabilistic classifiers:
\begin{itemize}
\item Samples with strong loyal type typically have much higher
predictive accuracy than the overall predictive accuracy.
\item Samples with weak loyal type typically have much lower
predictive accuracy than the overall predictive accuracy.
Therefore, one could refrain from using the predicted class
and switch to other means of classification instead.
\item LEM can be incorporated into the medical diagnosis problem to
minimize the total cost incurred in the diagnosis (see Section 3.3 below).
\item LEM can be more effectively
incorporated into larger machine learning tasks, in
a way that partially or completely avoids the problem of error propagation.
\end{itemize}

\subsection{Loyalty Type}
We will first generate $2m$ new classifiers
in designing the loyalty extraction machine.
To keep the explanation simple, we focus on
the case when $m=3$. Therefore, there are total
of six new classifiers. It is easy to see that
extension to the general $m$ is quite straightforward.

\medskip
Recall that the loss function $\phi$ is defined by
$$\phi=\phi_1+\phi_2+\phi_3$$ 
where $\phi_i$'s are defined by (\ref{s22-4}).
Let $\beta<1$ and $\gamma>1$ be positive numbers.
Consider the following six altered loss functions.
\begin{eqnarray}
\phi&=&\beta\phi_1+\phi_2+\phi_3   ~~~\Longrightarrow~~~\hat{\mu}(1,\beta,\cdot)
\label{s31-1} \\
\phi&=&\gamma\phi_1+\phi_2+\phi_3  ~~~\Longrightarrow~~~\hat{\mu}(1,\gamma,\cdot)
\label{s31-2} \\
\phi&=&\phi_1+\beta\phi_2+\phi_3   ~~~\Longrightarrow~~~\hat{\mu}(2,\beta,\cdot)
\label{s31-3} \\
\phi&=&\phi_1+\gamma\phi_2+\phi_3  ~~~\Longrightarrow~~~\hat{\mu}(2,\gamma,\cdot)
\label{s31-4} \\
\phi&=&\phi_1+\phi_2+\beta\phi_3   ~~~\Longrightarrow~~~\hat{\mu}(3,\beta,\cdot)
\label{s31-5} \\
\phi&=&\phi_1+\phi_2+\gamma\phi_3  ~~~\Longrightarrow~~~\hat{\mu}(3,\gamma,\cdot)
\label{s31-6}
\end{eqnarray}
For each of the altered loss functions, we carry out the classification task
in the similar fashion as given in Section 2 
and obtain a corresponding classifier
($\hat{\mu}(\cdot,\cdot,\cdot)$'s in (\ref{s31-1})-(\ref{s31-6})).
Hence six new classifiers are hatched.
These classifiers usually differ from the nominal classifier (that is, the one
with the unaltered loss function). They also differ from each other
in a way should become less ambiguous in a moment.
The rationale for doing so centers on differentiating the adherence of
samples pertaining to one member set from those pertaining to all other member sets. 

\medskip
We provide further explanation by using the loss function defined by (\ref{s31-1}).
Since $\beta<1$, it raises a situation
that, during the course of optimization, violation of the inequalities
\begin{equation}
f_1(x)<f_2(x),~f_1(x)<f_3(x) \label{s31-7}
\end{equation}
for any $x\in\Omega_1$ will be partially ignored.
Therefore, the optimization algorithm will place emphasis in
keeping other inequalities stay away from being violated. As a result,
this will decrease the chance for 1
to be selected as the predicted class for every training sample.
Given the circumstances, suppose that we have some $x\in\Omega_1$ whose predicted
class is still 1. Then it is an indication that $x$
has strong adherence to the class 1.

\medskip
Let us look at a different case when the loss function is defined by (\ref{s31-2}).
Since $\gamma>1$, it raises a situation
that, during the course of optimization, violation of the inequalities (\ref{s31-7})
for any $x\in\Omega_1$ will be more recognized.
Therefore, the optimization algorithm will place emphasis in
keeping (\ref{s31-7}) stay away from being violated. As a result, this
will increase the chance for 1 to be
selected as the predicted class for every training sample.
Similar statements can be made if the loss function is defined by
(\ref{s31-4}) or by (\ref{s31-6}).
Now suppose for some $x\in\Omega$ whose predicted class is 1 when
the loss function is defined by (\ref{s31-2}), is 2 
when the loss function is defined by (\ref{s31-4}), and is 3
when the loss function is defined by (\ref{s31-6}), that is, $x$
has a foot in both camps.
Given the circumstances, 
it indicates that $x$
has weak adherence to every class.

\medskip
In light of these observations, we 
can provide the definition for the loyalty type.
We reiterate that assigning a loyalty type to an unseen sample
is solely based on the training set. It has nothing
to do with other unseen samples.

\medskip\noindent
{\bf Definition 6.} A sample $x$ is said to be of {\it strong loyalty type}
if the set
$$\{i\in\M:\hat{\mu}(i,\beta,x)=i\}$$
is a singleton.
A sample $x$ is said to be of {\it weak loyalty type}
if it is not of strong loyalty type and the set
$$\{i\in\M:\hat{\mu}(i,\gamma,x)=i\}$$
has at least three elements.
A sample $x$ is said to be of {\it normal loyalty type}
if it is neither of strong nor of weak loyalty type.
\hfill$\BlackBox$

\medskip\noindent
It is easy to see that Definition 6 implies
every sample can only be either of strong or of normal loyalty type
for the case when $m=2$.

\subsection{Confusion Tensor}
A confusion tensor is a generalization of the confusion matrix to
adopt the loyalty type.
It is a three-dimensional matrix of size $3\times m\times m$.
Such an object has three layers,
each layer is a confusion matrix constructed for samples
corresponding to a loyalty type. Many performance measures such as
accuracy, precision and recall can be similarly defined with
respect to the confusion matrix in each layer.
\subsection{Medical Diagnosis Problem}
In this section, we introduce a mathematical problem which is called 
the {\it medical diagnosis problem}. Its formulation is made
possible by using the loyalty extraction machine.

\medskip
This paragraph uses excerpts from Wikipedia for wording. 
Medical diagnosis here is referred to the process of
determining whether a particular
disease or condition explains a person's symptoms and medical
signs. A symptom is a subjective or objective
departure from normal function or feeling which 
is apparent to a patient, reflecting the presence
of an unusual state, or of a disease.
A medical sign is an objective indication of some medical
fact or characteristic that may be detected by a patient
or anyone, especially a physician, before or
during a physical examination of a patient.
The information required for diagnosis is typically collected from
a history and physical examination of the person seeking
medical care. Often, one or more diagnostic procedure, such
as medical tests, are also done during the process.
A medical test is a medical procedure performed to detect,
diagnose, or monitor diseases, disease processes, susceptibility,
or to determine a course of treatment.
Medical tests relate to clinical chemistry and molecular
diagnostics, and are typically performed in a medical laboratory.

\medskip
Symptoms and medical signs is collectively called evidence.
The acquisition of an evidence may or may not involve a cost,
which may be either of monetary nature such as lab fees,
or intangible such as
waiting time or physical/psychological impact to the patient.
Let $\theta$ be the predictive accuracy of the medical diagnosis
derived from the full list of evidence
using the existing medical records (namely, the training set
in the context of machine learning).
Obviously, $\theta$ is the maximal predictive accuracy since, intuitively,
a medical diagnosis is likely to get worse using only a partial list
of evidence. When a new medical diagnosis process starts,
in order to reach predictive accuracy as high as possible,
it seems inevitable to acquire all evidence, and
thus results in a cost which may be unnecessarily high.
Remarkably and counterintuitively,
loyalty extraction machine demonstrates the ability of 
maintaining the maximal predictive accuracy,
while acquiring merely a partial list of
evidence in performing the medical diagnosis.
This observation makes the following problem a meaningful challenge.

\medskip\noindent
{\bf Medical Diagnosis Problem.} The problem is to determine the
optimal acquisition order of
evidence that satisfies the following two properties:
\begin{enumerate}
\item The acquisition process terminates
as soon as the evidence
collected so far are sufficient to determine whether
or not the patient has 
the disease or condition with the 
predictive accuracy no less than $\theta$.
\item The expected total diagnosis cost
is the lowest among all acquisition orders.
\hfill$\BlackBox$
\end{enumerate}

\section{Numerical Results}
In this section we give numerical results of running QMS and LEM. The experiment
runs on a notebook computer with specifications
given in Table \ref{Tab-s41-1}.
\begin{table}[htbp]
\centering
\small
\fbox{\parbox{7.7cm}{
\begin{tabular}{lr}
Model     & ThinkPad T14 Gen 1 \\
Memory    & 46.7 GiB \\
Processor & Intel Core i7 10510U @1.80GHz$\times$4 \\
OS Name   & Fedora 32 \\
OS Type   & 64-bit
\end{tabular}
}}
\caption{Computer specifications}
\label{Tab-s41-1}
\end{table}
Our code is implemented in the C language using single precision
arithmetic and SIMD instructions. The code only invokes a single user thread
and does not use any GPU.

\subsection{Dataset}
We use the Fashion MNIST dataset.
\begin{figure}[htbp]
\centerline{\includegraphics[width=2.5in,height=2.5in]{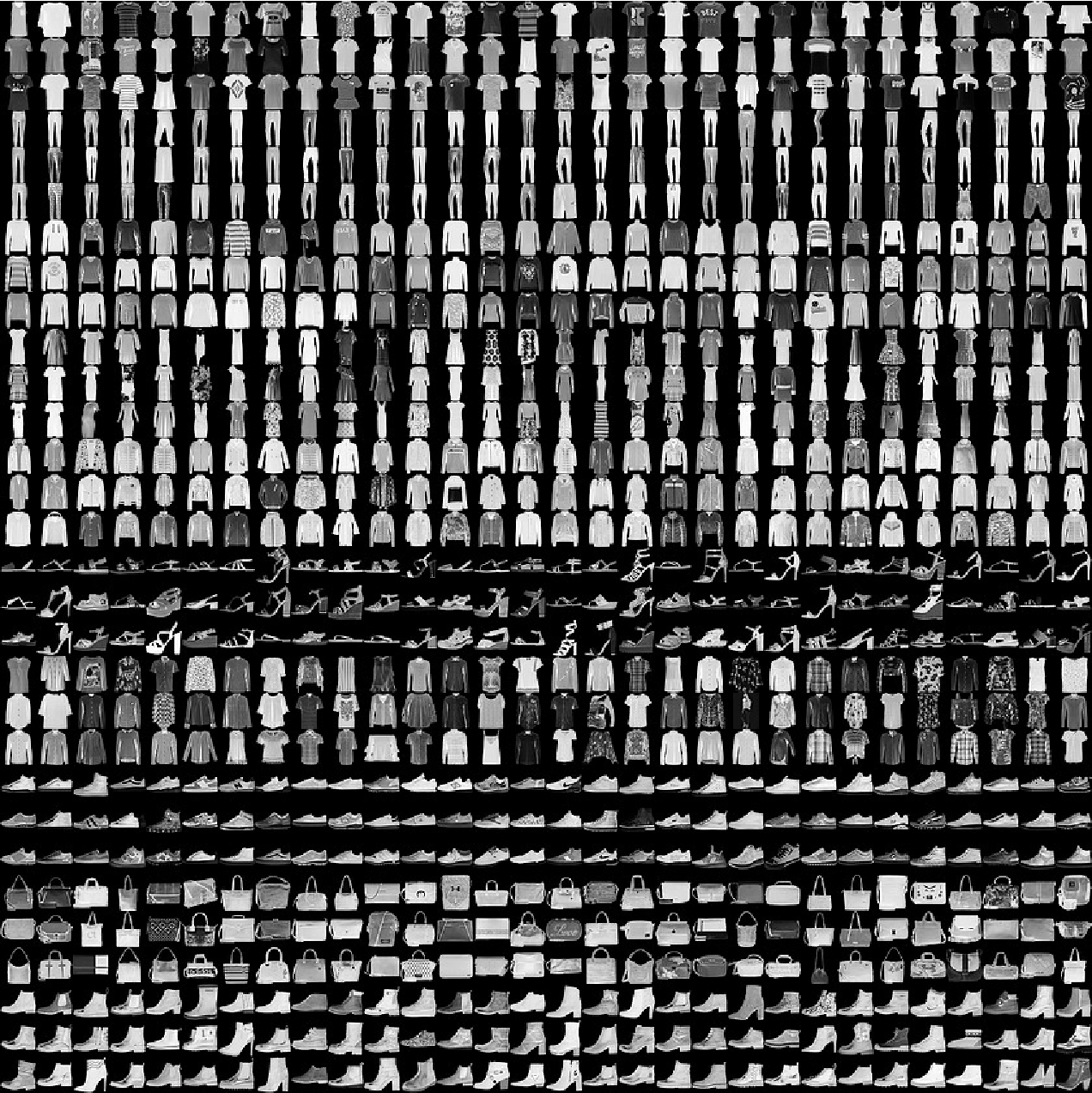}}
\caption{Fashion MNIST samples (by Zalando, MIT License)}
\end{figure}
It is a dataset of Zalando's article images, consisting
of a training set of 60,000 samples and a test set of
10,000 samples. Each sample is a 28x28 grayscale image,
associated with a label from 10 classes.
Fashion MNIST is intended to serve as a direct drop-in
replacement for the original MNIST dataset \citep{MNIST}.
The MNIST dataset
is often used as the "Hello, World" of 
benchmarking machine learning algorithms for computer vision. It
contains images of handwritten digits in a format
identical to that of the articles of clothing in the Fashion MNIST dataset.
Both datasets are relatively small and are used to verify
that an algorithm works as expected.
They are good starting points to test and debug code.

\subsection{Running QMS}
We follow the design procedure outlined in Section 2 and choose $\alpha=0.5$.
The classifier uses $q$-dimensional
member functions with $q=18$ for all separators.
Table \ref{Tab-s43-1} summarizes the result.
The resulting confusion matrices are depicted in Table
\ref{Tab-s43-2} and Table \ref{Tab-s43-3}.
\begin{table}[htbp]
\centering
\fbox{\parbox{7.5cm}{
\small
\begin{tabular}{lr}
Training accuracy & 92.69\% \\
Test accuracy & 88.63\% \\
Execution time & \hspace{1.17in}58 sec \\
\end{tabular}
}}
\caption{QMS result}
\label{Tab-s43-1}
\end{table}
\begin{table}[htbp]
\centering
\fbox{\parbox{11.3cm}{
\small
\begin{tabular}{rrrrrrrrrr}
   5422&    1&   61&  108&   14&    2&  375&    0&   16&    1 \\
      0& 5931&    1&   55&    4&    0&    6&    0&    2&    1 \\
     58&    1& 5198&   44&  399&    0&  290&    0&   10&    0 \\
     86&   10&   21& 5679&  124&    0&   77&    0&    3&    0 \\
     12&    3&  311&  127& 5290&    0&  249&    0&    7&    1 \\
      0&    0&    0&    0&    0& 5885&    0&   69&    3&   43 \\
    543&    2&  368&   95&  318&    0& 4646&    0&   27&    1 \\
      0&    0&    0&    0&    0&   53&    0& 5830&    0&  117 \\
      9&    0&    8&   17&   10&    4&   19&    6& 5922&    5 \\
      0&    0&    0&    0&    1&   35&    0&  148&    3& 5813
\end{tabular}
}}
\caption{Confusion matrix (training set)}
\label{Tab-s43-2}
\end{table}
\begin{table}[htbp]
\centering
\fbox{\parbox{11.3cm}{
\small
\begin{tabular}{rrrrrrrrrr}
    858&    0&   14&   26&    4&    1&   91&    0&    6&    0 \\
      4&  963&    5&   19&    7&    0&    2&    0&    0&    0 \\
     21&    1&  814&   10&   88&    0&   63&    0&    3&    0 \\
     25&    4&   12&  897&   31&    0&   25&    0&    6&    0 \\
      1&    1&   79&   27&  830&    0&   62&    0&    0&    0 \\
      0&    0&    0&    1&    0&  949&    0&   19&    2&   29 \\
    141&    1&   83&   27&   66&    0&  674&    0&    8&    0 \\
      0&    0&    0&    0&    0&   18&    0&  954&    0&   28 \\
      3&    0&    8&    6&    6&    3&   11&    3&  960&    0 \\
\hspace{0.204in}0&
\hspace{0.204in}0&
\hspace{0.204in}0&
\hspace{0.204in}0&
\hspace{0.204in}0&
\hspace{0.204in}5&
\hspace{0.204in}1&
\hspace{0.15in}30&
\hspace{0.204in}0&
\hspace{0.07in}964
\end{tabular}
}}
\caption{Confusion matrix (test set)}
\label{Tab-s43-3}
\end{table}

\subsection{Running LEM}
We follow the design procedure outlined in Section 3 and
choose $\beta=0.05$ and $\gamma=20$. Twenty additional classifiers are constructed.
Table \ref{Tab-s44-1} summarizes the result for the training set.
In Table \ref{Tab-s44-1}, $n$ denotes the total number of samples (which is 60000).
For each loyalty type, $n_1$ is the number of samples of that type and
$n_2$ is the number of samples with correct predictive class.
Also, $n_2/n_1$ is the {\it local} predictive accuracy (LPA).
The overall predictive accuracy 92.69\% can also be computed either by adding
all numbers in column $n_2$ and then divided by $n$, or
by taking the inner product of the two columns $n_1/n$ and LPA.
\begin{table}[htbp]
\centering
\fbox{\parbox{9.6cm}{
\begin{tabular}{lrrrr}
loyalty type & $n_1$ & $n_1/n$ & $n_2$ & $n_2/n_1$ \\
strong&\small 43,249&\small 72.08\%&\small 43,173&\small 99.82\% \\
normal&\small 13,118&\small 21.86\%&\small 10,473&\small 79.84\% \\
weak  &\small  3,633&\small  6.06\%&\small  1,970&\small 54.23\% \\
\vspace{-0.2in}
\hspace{0.53in}~&
\hspace{0.53in}~&
\hspace{0.53in}~&
\hspace{0.53in}~&
\hspace{0.53in}~
\end{tabular}
}}
\caption{LEM result (training set)}
\label{Tab-s44-1}
\end{table}

\medskip\noindent
Table \ref{Tab-s44-2} summarizes the result for the test set,
where $n$ here is equal to 10000.
\begin{table}[htbp]
\centering
\fbox{\parbox{9.6cm}{
\begin{tabular}{lrrrr}
loyalty type & $n_1$ & $n_1/n$ & $n_2$ & $n_2/n_1$ \\
strong&\small 6,967&\small 69.67\%&\small 6,789&\small 97.45\% \\
normal&\small 2,299&\small 22.99\%&\small 1,699&\small 73.90\% \\
weak  &\small   734&\small  7.34\%&\small   375&\small 51.09\% \\
\vspace{-0.2in}
\hspace{0.53in}~&
\hspace{0.53in}~&
\hspace{0.53in}~&
\hspace{0.53in}~&
\hspace{0.53in}~
\end{tabular}
}}
\caption{LEM result (test set)}
\label{Tab-s44-2}
\end{table}

\medskip
Let us give the highlights:
\begin{itemize}
\item For the test set, 69.67 percent of samples are of strong
loyalty type.
\item For any samples of strong loyalty type,
the {\it estimated} predictive accuracy is 99.82 percent.
\item For samples of strong loyalty type in the test set,
the {\it actual} predictive accuracy is 97.45 percent.
\end{itemize}

\medskip
Finally, layers of the confusion tensor are depicted in Table \ref{Tab-s44-3},
Table \ref{Tab-s44-4}, and Table \ref{Tab-s44-5}, respectively.
Notice that sum of those layers is equal to the confusion matrix given
in Table \ref{Tab-s43-3}.
\begin{table}[htbp]
\centering
\fbox{\parbox{9.6cm}{
\small
\begin{tabular}{rrrrrrrrrr}
    530&     0&     2&     4&     0&     0&     9&     0&     2&     0 \\
      0&   943&     2&    11&     1&     0&     0&     0&     0&     0 \\
      1&     0&   456&     5&     5&     0&     1&     0&     0&     0 \\
      3&     1&     2&   691&     2&     0&     3&     0&     2&     0 \\
      0&     1&     8&     3&   406&     0&     8&     0&     0&     0 \\
      0&     0&     0&     0&     0&   882&     0&     7&     0&    14 \\
     21&     0&    12&     5&    10&     0&   256&     0&     5&     0 \\
      0&     0&     0&     0&     0&     2&     0&   824&     0&     8 \\
      0&     0&     2&     2&     2&     1&     2&     0&   925&     0 \\
      0&     0&     0&     0&     0&     2&     1&     6&     0&   876 \\
\end{tabular}
}}
\caption{Confusion tensor, strong loyalty type layer (test set)}
\label{Tab-s44-3}
\end{table}
\begin{table}[htbp]
\centering
\fbox{\parbox{9.6cm}{
\small
\begin{tabular}{rrrrrrrrrr}
    286&     0&     5&    12&     1&     0&    61&     0&     3&     0 \\
      4&    14&     3&     6&     2&     0&     2&     0&     0&     0 \\
      5&     0&   297&     1&    58&     0&    30&     0&     0&     0 \\
      9&     2&     3&   154&    19&     0&     9&     0&     3&     0 \\
      1&     0&    47&    18&   329&     0&    25&     0&     0&     0 \\
      0&     0&     0&     1&     0&    64&     0&    11&     2&    14 \\
     93&     0&    39&     8&    29&     0&   319&     0&     2&     0 \\
      0&     0&     0&     0&     0&    16&     0&   129&     0&    18 \\
      1&     0&     3&     2&     0&     2&     3&     3&    21&     0 \\
\hspace{0.138in}0&
\hspace{0.138in}0&
\hspace{0.138in}0&
\hspace{0.138in}0&
\hspace{0.138in}0&
\hspace{0.138in}3&
\hspace{0.138in}0&
\hspace{0.067in}21&
\hspace{0.138in}0&
\hspace{0.067in}86 
\end{tabular}
}}
\caption{Confusion tensor, normal loyalty type layer (test set)}
\label{Tab-s44-4}
\end{table}
\begin{table}[htbp]
\centering
\fbox{\parbox{9.6cm}{
\small
\begin{tabular}{rrrrrrrrrr}
     42&     0&     7&    10&     3&     1&    21&     0&     1&     0 \\
      0&     6&     0&     2&     4&     0&     0&     0&     0&     0 \\
     15&     1&    61&     4&    25&     0&    32&     0&     3&     0 \\
     13&     1&     7&    52&    10&     0&    13&     0&     1&     0 \\
      0&     0&    24&     6&    95&     0&    29&     0&     0&     0 \\
      0&     0&     0&     0&     0&     3&     0&     1&     0&     1 \\
     27&     1&    32&    14&    27&     0&    99&     0&     1&     0 \\
      0&     0&     0&     0&     0&     0&     0&     1&     0&     2 \\
      2&     0&     3&     2&     4&     0&     6&     0&    14&     0 \\
\hspace{0.138in}0&
\hspace{0.138in}0&  
\hspace{0.138in}0& 
\hspace{0.138in}0& 
\hspace{0.138in}0& 
\hspace{0.138in}0& 
\hspace{0.138in}0&
\hspace{0.138in}3&  
\hspace{0.138in}0& 
\hspace{0.138in}2 \\
\end{tabular}
}}
\caption{Confusion tensor, weak loyalty type layer (test set)}
\label{Tab-s44-5}
\end{table}

\section*{Appendix A. Proofs}
{\bf Proof of Lemma 1}. 
A member function is certainly nonnegative. Let 
\begin{equation*}
f(x)=\sum_{1\le i\le j \le p} a_{i j}x_i x_j 
+ \sum_{i=1}^p a_i x_i +a_0
\end{equation*}
be a real, nonnegative quadratic polynomial. 
The way to express it as a member function is not unique.
We describe one possible way below.

Clearly, $a_0\ge 0$ and $a_{ii}\ge 0$ for all $i$ since $f$ is nonnegative.
Moreover, if $a_{ii}=0$ for some $i$, then $a_i=a_{ij}=0,
j\ne i$, since otherwise one could choose 
suitable constants $c_j, j\ne i$, to obtain
a nontrival linear function $f(c_1, \dots, c_{i-1}, x_i, c_{i+1}, \dots,  c_p)$
which cannot be nonnegative. 
Thus, without
loss of generality, we may simply consider the case 
$a_{ii}>0, i=1, \dots, p$. 

Define $f_j, g_j, j=1, \dots, p$, inductively as follows.
For $j=1$, define $f_1=f-g_1$, and 
\[ 
g_1(x)= L_1(x)^2,
\]
where $L_1(x)$ is a linear polynomial in variables
$x_1, \dots, x_p$ :
\begin{equation*}
	L_1(x)=\sqrt{a_{1 1}} x_1+
	\frac  {a_{1 2}}{2\sqrt{a_{1 1}}}x_2+\cdots
	+ \frac  {a_{1 p}}{2\sqrt{a_{1 1}}}x_p+
	\frac{a_1}{2\sqrt{a_{1 1}}} .
\end{equation*}

Since $g_1$ collects all the terms in $f$ containing variable $x_1$, 
the function
\begin{eqnarray*}
	f_1(x) &\equiv& f(x)-g_1(x)=
	\left(a_{2 2}- \frac  {a^2_{1 2}}{4 a_{1 1}} \right)
	x_2^2+\cdots +\left(a_0-\frac{a_1^2}{4a_{1 1}}\right) \\
	&\equiv& 	\sum_{2\le i \le j\le p} a^{(1)}_{i j}x_i x_j + 
	\sum_{2\le i\le p} a^{(1)}_i x_i +a^{(1)}_0
\end{eqnarray*}
becomes a polynomial without variable $x_1$.
We claim that 
$f_1$ is also nonnegative.
Suppose not, then 
$f_1(\cdot, \tilde t_2, \dots, \tilde t_p)<0$ for some
$(\tilde t_2, \dots, \tilde t_p)\in \mathbb{R}^{p-1}$.
Now solve $\tilde t_1\in \mathbb R$ such that 
the linear equation $L_1(\tilde t)=0$,
where $\tilde t= (\tilde t_1, \dots, \tilde t_p)$. 
Given $(\tilde t_2, \dots, \tilde t_p), \tilde t_1$ can be solved since $a_{11}>0$. Plugging $\tilde t$ into $f$ then
yields a contradiction $f(\tilde t)= f_1(\tilde t)+g_1(\tilde t)
=f_1(\tilde t)+0 <0$.

The nonnegativeness of $f_1$ implies
\[
a^{(1)}_{2 2}=a_{2 2}- \frac  {a^2_{1 2}}{4 a_{1 1}} \ge 0, 
\dots, a^{(1)}_{p\, p} \ge 0, 
a_0^{(1)}=a_0-\frac{a_1^2}{4a_{1 1}} \ge 0 .
\]
If all $a^{(1)}_{ii}=0, i=2, \dots, p$, we are done.
Otherwise, assume $a^{(1)}_{2 2}>0$ with possibly re-indexing
the variables.
Let $g_2=L_2(x)^2$ and $f_2=f_1- g_2$, where
$L_2$ is a linear polynomial in variables $x_2, \dots, x_p$ :
\begin{equation*}
	L_2(x)=\sqrt{a^{(1)}_{2 2}} x_2+
	\frac  {a^{(1)}_{2 3}}{2\sqrt{a^{(1)}_{2 2}}}x_3+\cdots
	+ \frac  {a^{(1)}_{2 p}}{2\sqrt{a^{(1)}_{2 2}}}x_p+
	\frac{a^{(1)}_2}{2\sqrt{a^{(1)}_{2 2}}} .
\end{equation*}

Since $g_2$ collects all the terms
in $f_1$ containing variable $x_2$, 
$f_2=f_1- g_2$ is a polynomial in variables $x_3, \dots, x_p$.
The nonnegativeness of $f_2$ can be seen easily.
However, for simplicity, below we give a more
general observation.

Suppose that the decomposition process described above has been carried
out $k, k\le p$, times and we reach the expression
$f=g_1+g_2+\cdots+g_k+f_k$,
where $g_j(x)=L_j(x)^2$ and $L_j(x)=\sqrt{a^{(j-1)}_{j j}} x_j+\cdots, 
a^{(j-1)}_{j j}>0, j=1, \dots, k$,  
is a linear polynomial in variables $x_j, x_{j+1}, \dots, x_p$.
This process ensures that $f_k$ is a quadratic polynomial independent of $x_1, \dots, x_k$.
We claim that $f_k$ is nonnegative so that the induction argument can proceed.
If not, there exists $(\tilde t_{k+1}, \dots, \tilde t_p)\in \mathbb R^{p-k}$
such that 
$f_k(\cdot, \tilde t_{k+1}, \dots, \tilde t_p)<0$.
By the fact $k\le p$ and the upper triangular nature of the 
system of $k$ linear equations,
$k$ unknowns $\tilde t_{1}, \dots, \tilde t_k \in \mathbb R$ can be found
so that 
$L_1(\tilde t) = \cdots = L_k(\tilde t) =0$,
where $\tilde t= (\tilde t_1, \dots, \tilde t_p)$.
Once more this gives a contradiction $f(\tilde t)=0+
f_k(\tilde t) < 0$. Hence $f_k$ must be nonnegative.

In the end the decomposition process stops
at some expression $f_{\ell-1}=g_\ell + f_\ell$, where
$f_\ell$ is nonegative and independ of all variables $x_1, \dots, x_p$.
Consequently, $f_\ell$ must be a nonnegative constant $a^{(\ell)}_0\ge 0$.
Obviously, the number $\ell$ of decompositions
must be less than or equal to $p$, the number of variables. 
The decomposition 
\begin{equation*}
f=g_1+\cdots +g_\ell+a^{(\ell)}_0
\end{equation*}
indicates how the entries of both
$A\in \mathbb R^{q\times p}$ and $b\in \mathbb R^q$ 
can be chosen with $q \le \ell +1 \le p+1$. This completes the proof.

\hfill\BlackBox

\medskip\noindent
{\bf Proof of Lemma 2}. 
In view of Lemma 1,
it suffices 
to prove that every quadratic polynomial can be written
as a difference of two nonnegative quadratic polynomials.
The case of monomials can be easily checked.
For example, 
$x_ix_j=\frac 14 \left[ (x_i+x_j)^2 - (x_i-x_j)^2 \right]$, 
$x_i= \frac 14 \left[ (x_i+ 1)^2 - (x_i - 1)^2 \right]$.
The general case follows trivially. 
\hfill\BlackBox

\medskip\noindent
{\bf Proof of Theorem 1.} By construction.
It suffices to show that there exist member functions $f_1,\dots,f_m$ such that
\begin{equation}
h_{ij}=f_i-f_j {\rm~~for~all~}i,j\in\M \label{sa-1}
\end{equation}
First it is easy to see that the cyclic conditions imply
$h_{ii}=0$ for all $i\in\M$, and $h_{ji}=-h_{ij}$ for all $i,j\in\M$.
In view of Lemma 2, for $i\in\M$,
there exist member functions $u_i$ and $v_i$ such that
$$h_{1i}=u_i-v_i$$
Notice that $u_1=v_1$. For $i\in\M$, define the member function $f_i$ by
$$f_i=v_i-u_i+\sum_{k=1}^mu_k$$
Consequently we have $f_1=\sum_{k=1}^mu_k$ and
$f_i=v_i-u_i+f_1$ for all $i\in\M$.
Finally, the validity of (\ref{sa-1})
follows from direct verification as follows.
$$h_{ij}=h_{1j}-h_{1i}=(u_j-v_j)-(u_i-v_i)=(f_1-f_j)-(f_1-f_i)=f_i-f_j$$
The proof is then complete.\hfill\BlackBox

\bibliography{Reference}
\end{document}